\documentclass{article}

\usepackage{tikz}
\usepackage{arxiv}
\usepackage[utf8]{inputenc} % allow utf-8 input
\usepackage[T1]{fontenc}    % use 8-bit T1 fonts
\usepackage[hidelinks]{hyperref}       % hyperlinks
\usepackage{url}            % simple URL typesetting
\usepackage{booktabs}       % professional-quality tables
\usepackage{amsfonts}       % blackboard math symbols
\usepackage{nicefrac}       % compact symbols for 1/2, etc.
\usepackage{microtype}      % microtypography
\usepackage{lipsum}		% Can be removed after putting your text content
\usepackage{graphicx}
\usepackage{natbib}
\usepackage{doi}
\usepackage{multirow}
\usepackage{amsmath}
\usepackage{float}
\usepackage{subcaption}

\title{A strong inductive bias: Gzip for binary image classification}

\author{ {Marco Scilipoti} \\
	Department of Computer Engineering\\
	Instituto Tecnológico de Buenos Aires (ITBA)\\
	Buenos Aires, Argentina \\
	\texttt{mscilipoti@itba.edu.ar} \\
	\And  
 	{Marina Fuster} \\
	Department of Computer Engineering\\
	Instituto Tecnológico de Buenos Aires (ITBA)\\
	Buenos Aires, Argentina \\
	\texttt{mfuster@itba.edu.ar} \\
    \And
	\href{https://orcid.org/0000-0001-8155-0124}{\includegraphics[scale=0.06]{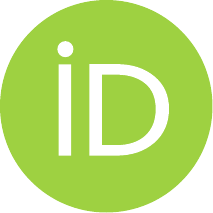}\hspace{1mm} Rodrigo Ramele} \\
	Department of Computer Engineering\\
	Instituto Tecnológico de Buenos Aires (ITBA)\\
	Buenos Aires, Argentina \\
	\texttt{rramele@itba.edu.ar} \\
}

\renewcommand{\undertitle}{}
\renewcommand{\shorttitle}{A strong inductive bias: Gzip for binary image classification}

\hypersetup{
pdftitle={A strong inductive bias: Gzip for binary image classification},
pdfsubject={q-bio.NC, q-bio.QM},
pdfauthor={Marco Scilipoti, Rodrigo Ramele},
pdfkeywords={inductive bias,CV,parameter-less,compressor},
}

\begin{document}
\usetikzlibrary{positioning, arrows.meta, shapes}

\maketitle

\begin{abstract}
Deep learning networks have become the de-facto standard in Computer Vision for industry and research. However, recent developments in their cousin, Natural Language Processing (NLP), have shown that there are areas where parameter-less models with strong inductive biases can serve as computationally cheaper and simpler alternatives. We propose such a model for binary image classification: a nearest neighbor classifier combined with a general purpose compressor like Gzip. We test and compare it against popular deep learning networks like Resnet, EfficientNet and Mobilenet and show that it achieves better accuracy and utilizes significantly less space—more than two orders of magnitude—within a few-shot setting. As a result, we believe that this underlines the untapped potential of models with stronger inductive biases in few-shot scenarios. 
\end{abstract}

% keywords can be removed

\section{Introduction}

Inductive bias describes the preference for solutions that a machine learning algorithm holds before seeing any data. It is an intrinsic component, not often made explicitly, for the goal of machine learning, which is to generalize from a set of examples to unseen data points, and it is encoded in the underlying architecture or mechanics of several deep learning models~\citep{rendsburg2023inductive}.

Convolutional neural networks (CNNs) have been foundational in image processing since their inception by Fukushima and LeCun  \citep{fukushima1989analysis,lecun1989backpropagation} and their re-introduction in AlexNet \citep{AlexNet}. The architectural components of CNNs, such as local connectivity and weight sharing, have been cornerstone to their success. These inductive biases, particularly spatial equivariance, ensure that CNNs can effectively detect patterns regardless of their spatial location, an essential feature for visual data \citep{GoodBengCour16}. 

More recently, transformer models have emerged as a powerful tool for various tasks, including image classification. The Vision Transformer (ViT) \citep{ViT} epitomizes this shift. Unlike CNNs, which have biases hard-coded into their architecture, the authors remark how transformers rely on data to derive essential features. For instance, \citep{ViT} illustrates the principal components of the initial RGB layer in values of the ViT-L/32 network, resembling CNN kernels as shown in \citep{zeiler2013visualizing}. However, this adaptability also means transformers often require more data to achieve similar performance levels as CNNs.

Because such an architecture with a weaker inductive bias requires more data to generalize, we hypothesize that an architecture with a stronger inductive bias will generalize better in a few-shot setting.  As a result, we propose a parameter-less model with a strong inductive bias for image classification and compare it to the several deep learning architectures. There has been some research done on the inductive biases like in \citep{battaglia2018relational} with the formalization of the graph-network, as well as a rapport between inductive biases and higher level cognition by \citep{goyal2022inductive}. We seek to open the playing field and introduce a non deep-learning model with a strong inductive bias.

Drawing from similar work done in natural language processing by  \citep{jiangetal2023low}, the architecture of the proposed model is as follows: a general compressor, a nearest neighbor classifier and \textit{Normalized Compression Distance} (NCD) as defined by \citep{li2004similarity}. 
% Moreover, we present a more efficient extension based on the \textit{compression similarity measurement} (CSM) \citep{keog2004disimilarity}. 

Our aim is to highlight the untapped potential of architectures with stronger inductive biases in terms of both accuracy and computational efficiency. The main contributions of the paper can be summarized as follows:
\begin{itemize}
\item Introducing Gik (Gzip Image kNN), a parameter-less model featuring a strong inductive bias that exhibits superior accuracy compared to conventional deep learning models in a few-shot learning setting
\item Demonstrating that this architecture requires substantially less computational space than pre-trained deep learning models designed for running on mobile devices and memory constrained environments.
\end{itemize}

This paper proceeds as follows: first we introduce in Section \ref{Gik} the \textbf{Gik} architecture and describe the deep learning models that we are using for comparison. In Section \ref{datasets} the two datasets that were used for the experiments are described. Results are presented in the Section \ref{results}  and conclusion and future work are presented in the last Section~\ref{conclusion}.

\section{Method and Materials}

\subsection{Normalized Compression Distance}
Kolmogorov complexity of an input \( x \) is defined as: \textit{the length of the shortest binary program that can describe \( x \)} \citep{Kolmogorov1998OnTO}. Similarly, we can define the conditional Kolmogorov complexity as,

\begin{equation}
K(y|x) = K(x,y) - K(x).
\end{equation}

Intuitively, it can be understood as the length of the shortest program that encodes the concatenation of inputs \( x \) and \( y \), minus the complexity of \( x \).  With the idea of conditional Kolmogorov complexity, \citep{bennett2010information} defined \textit{information distance} as,
\begin{equation}
E(x, y) = \max\{K(x|y), K(y|x)\}
\end{equation}
\noindent which can be understood as the length of the shortest binary program that computes \( x \) from \( y \) and vice versa. Because Kolmogorov complexity is uncomputable (and as a result, the information distance is as well), an approximation is required.  Proposed by \citep{li2004similarity}, \textit{normalized compression distance} (NCD) can be used in exchange of Kolmogorov information distance, by replacing "the length of the shortest program to reproduce \( x \)" with "the length in bits of the lossless compression of \( x \)". The compressed version can be decompressed losslessly, obtaining the original \( x \), thus fully describing it.

Mathematically, NCD is defined as follows:
\begin{equation}
\text{NCD}(x, y) = \frac{C(xy) - \min\{C(x), C(y)\}}{\max\{C(x), C(y)\}}
\end{equation}
\noindent where \( C(x) \) represents the length of the compressed version of \( x \), and \( C(xy) \) represents the length of the compression of \( x \) concatenated with \( y \).

% Extending this idea, we propose \textit{compression similarity distance} (CSM). It's based on the \textit{compression dissimilarity measurement} (CDM) defined by \citep{keog2004disimilarity} for data mining. Mathematically, it's described as,
% \begin{equation}
% \text{CSM}(x, y) = \frac{C(x) + C(y)}{C(xy)} - 1.
% \label{eq:csm}
% \end{equation}

% The intuition behind it can be explained by understanding how it behaves at the two extremes of the input space: (a) if \( x \) and \( y \) have no shared structure, then \( C(xy) = C(x) + C(y) \) because the compressor cannot generalize anything from them to encode their concatenation in fewer bits. Hence, if one substitutes this in equation~\ref{eq:csm} then \( CSM(x,y)= 0 \), and the inputs are completely dissimilar. On the other hand, (b) if \( x \) and \( y \) share all their structure, then \( C(xy) = C(x) = C(y) \), the compressor is able to encode all the structure of \( x \) with the information of \( y \). Symbolically, if one again substitutes this in equation~\ref{eq:csm} then \( CSM(x,y) = 1 \), which means that the inputs are completely similar.

\subsection{Gik:  Gzip Image Knn}
\label{Gik}
The architecture proposed is straightforward and comprises two main components: (1) a lossless compression algorithm and (2) a kNN classifier \cite{mccann2011local}. The lossless compression algorithm is gzip as defined by \cite{gzip}. This is used to compute the Normalized Compression Distance between a pair of images. The NCD gives us a scalar encoding for similarity between two images that is then used by kNN classifier to infer the category of the image.

\begin{figure*}[]
    \centering
   \begin{tikzpicture}[
    node distance=1cm and 2.5cm,
    auto,
    block/.style={rectangle, draw, fill=gray!20, align=center, minimum height=1cm, text width=2.5cm},
    arrow/.style={-Latex},
    line/.style={draw, -Latex}
]

    % Coordinates for image stack
    \coordinate (imageStart);
    \coordinate[right=1cm of imageStart] (encodingStart);

    % Stacked images
    \foreach \i in {0,...,4} {
        \node[inner sep=0, xshift=\i*0.1cm, yshift=-\i*0.1cm] at (imageStart) {\includegraphics[width=.1\textwidth]{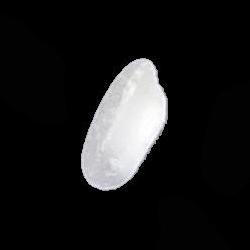}};
    }

    % Blocks
    \node [block, right=of encodingStart] (encoding) {1d encodings};
    \node [block, right=of encoding] (category) {categories};

    % Arrows with labels
    \draw [line] (encodingStart) -- (encoding) node[midway, above] {gzip + ncd};
    \draw [line] (encoding) -- (category) node[midway, above] {kNN};

\end{tikzpicture}
    \caption{Diagram of gik architecture}
    \label{fig:gik_diagram}
\end{figure*}
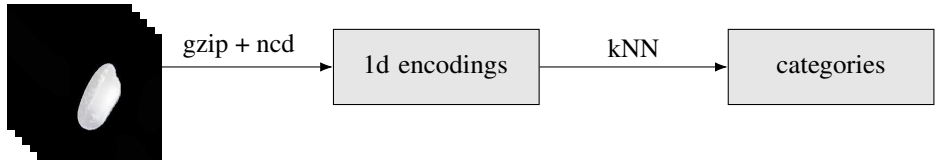

In \textit{Figure \ref{fig:gik_diagram}} we can see the inference sequence. For every training image we compute the normalized compression distance between that image and the test image. After computing such similarity encodings across all training images the nearest neighbor classifier subsequently employs these to locate the test image's closest neighbors and select the appropriate label. 

Although this approach can become computationally expensive, with its \(O(n^2)\) complexity where \(n\) is the training set size, it remains manageable in few-shot learning scenarios.

\subsection{Deep Learning Models}

For comparison with deep learning models, we selected three mainstream architectures: (a) ResNet18 (b) MobileNetV3-Small (c) EfficientNetB0. All models were pre-trained using the \textit{timm} library \citep{timm}. The rationale behind this choice is to contrast our proposed model with prevalent models designed for memory-constrained environments, examining both accuracy and space requirements. 

\begin{itemize}
\item ResNet \citep{he2015deep}. This network was chosen as a general baseline. Although not designed specifically for efficiency, it's widely used across the computer vision community and the ResNet18 is a great baseline for computer vision tasks. It has 11 million parameters
\item MobileNetV3 \citep{howard2019searching} This architecture family is characterized by the introduction of depth-wise separable convolutions. This allowed for considerable reductions in compute and space requirements. Nonetheless the version MobileNetv3-Small still has 2.5 million parameters.
\item EfficientNet \citep{tan2020efficientnet} This architecture family was designed by systematically optimizing depth, width and resolution through multi-objetive neural architecture \cite{tan2019mnasnet} search and the introduction of a compound scaling method. The smallest model, EfficientNetB0, has 5.3 million parameters.
\end{itemize}

\subsection{Datasets}
\label{datasets}
\begin{figure}[H]
    \begin{subfigure}[b]{0.2\textwidth}
        \includegraphics[width=\textwidth]{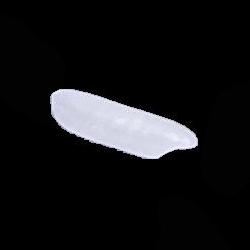}
        \caption{Jasmine class}
    \end{subfigure}
    \hfill
    \begin{subfigure}[b]{0.2\textwidth}
        \includegraphics[width=\textwidth]{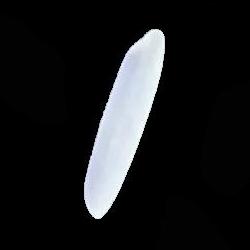}
        \caption{Basmati class}
    \end{subfigure}
    \hfill
    \begin{subfigure}[b]{0.2\textwidth}
        \includegraphics[width=\textwidth]{Images/Arborio.jpg}
        \caption{Arborio class}
    \end{subfigure}
    \hfill
    \begin{subfigure}[b]{0.2\textwidth}
        \includegraphics[width=\textwidth]{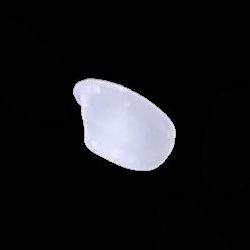}
        \caption{Karacadag class}
    \end{subfigure}
    \caption{Sample of the rice categories tested.}
    \label{fig:sample}
\end{figure}

Two datasets were created based on the \citep{riceDataset} rice dataset. Given our focus on binary classification, we sampled four of the original dataset's categories, thereby creating two distinct datasets: (a) Jasmine and Basmati and (b) Arborio and Karacadag. Refer to \textit{Figure \ref{fig:sample}} for sample images from each class. To facilitate a comparison of parameter-less models within a few-shot setting, we randomly selected 80 images from each category. This yielded a combined training and test sample size of 160 images for each dataset.

For the learning phase of the comparative models, all deep learning models were trained with a uniform configuration. Each image was resized to \(224 \times 224\) pixels, and their pixel values were normalized using z-scores from the ImageNet dataset \citep{imagenet}. All the models, pre-trained and defined within the \textit{timm} library \citep{timm}, were trained using early stopping, with a patience of 3 and a mean delta of 0.01, monitoring the validation loss. Furthermore, a maximum of 30 epochs was set for model fine tuning on the respective dataset, utilizing a learning rate of \(2 \times 10^{-3}\) and the Adam optimizer \citep{kingma2017adam}. The loss function used was cross-entropy loss.

The training process for the parameter-less models was more straightforward. After experimenting with different configuration of sizes, we settled for resizing all images to \(32 \times 32\) pixels and converted them to grayscale. 

Finally, regarding the training and test split, we adopted a sliding scale that ranged from a 90:10 to a 10:90 training-to-test ratio. This strategy enabled us to probe the models' effectiveness in few-shot scenarios while maximizing the test set's size. By adjusting the size of the training set in this way, we aimed to evaluate the models' capacity to generalize with limited training data more effectively.

\section{Results and Discussion}
\label{results}

\begin{figure*}[!tb]
    \centering
    \includegraphics[width=\textwidth]{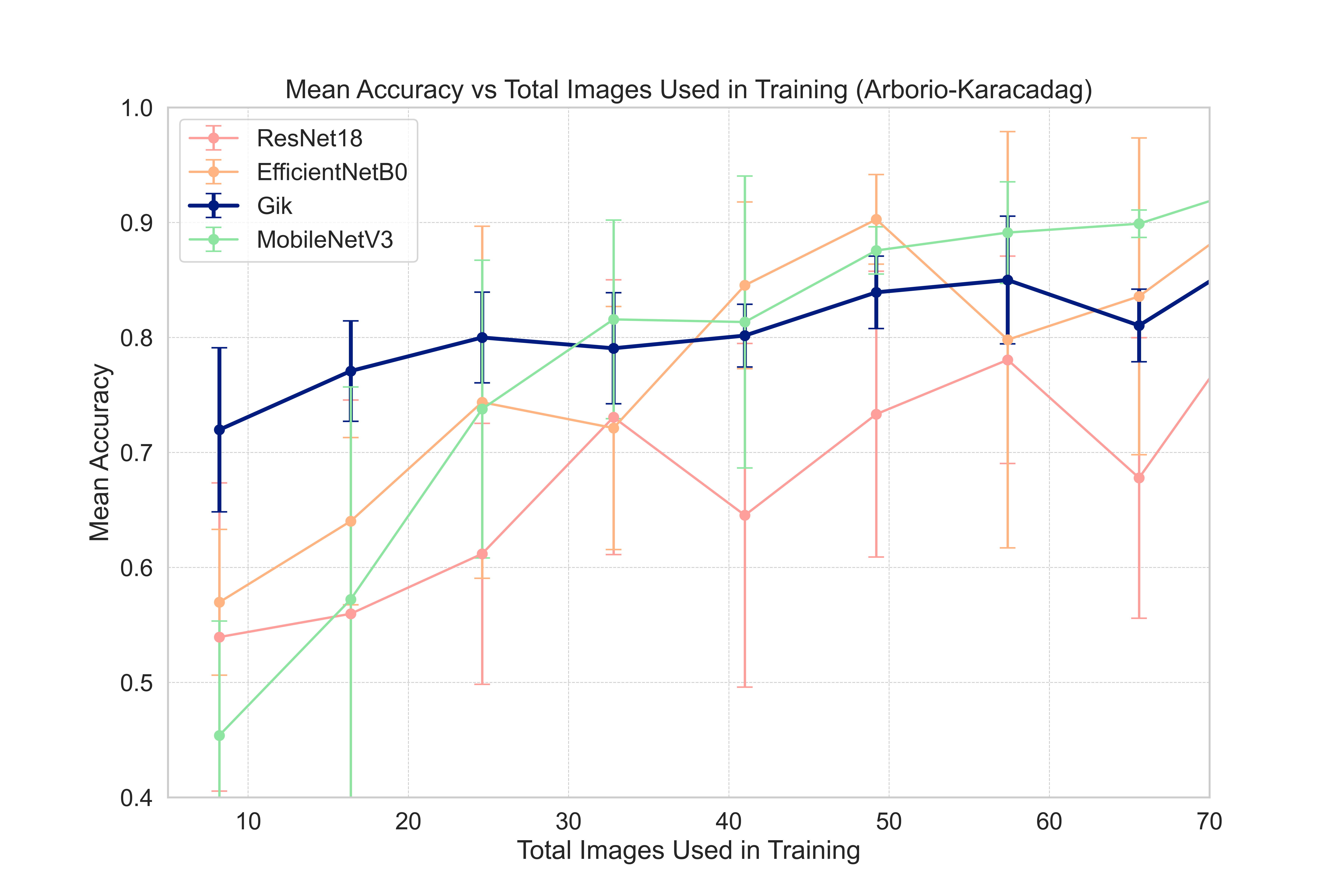}
    \caption{Mean accuracy in binary classification across the classes \textit{Jasmine} and \textit{Basmati}}
    \label{fig:JasmineBasmatiGraph}
\end{figure*}

\begin{figure*}[!t]
    \centering
    \includegraphics[width=\textwidth]{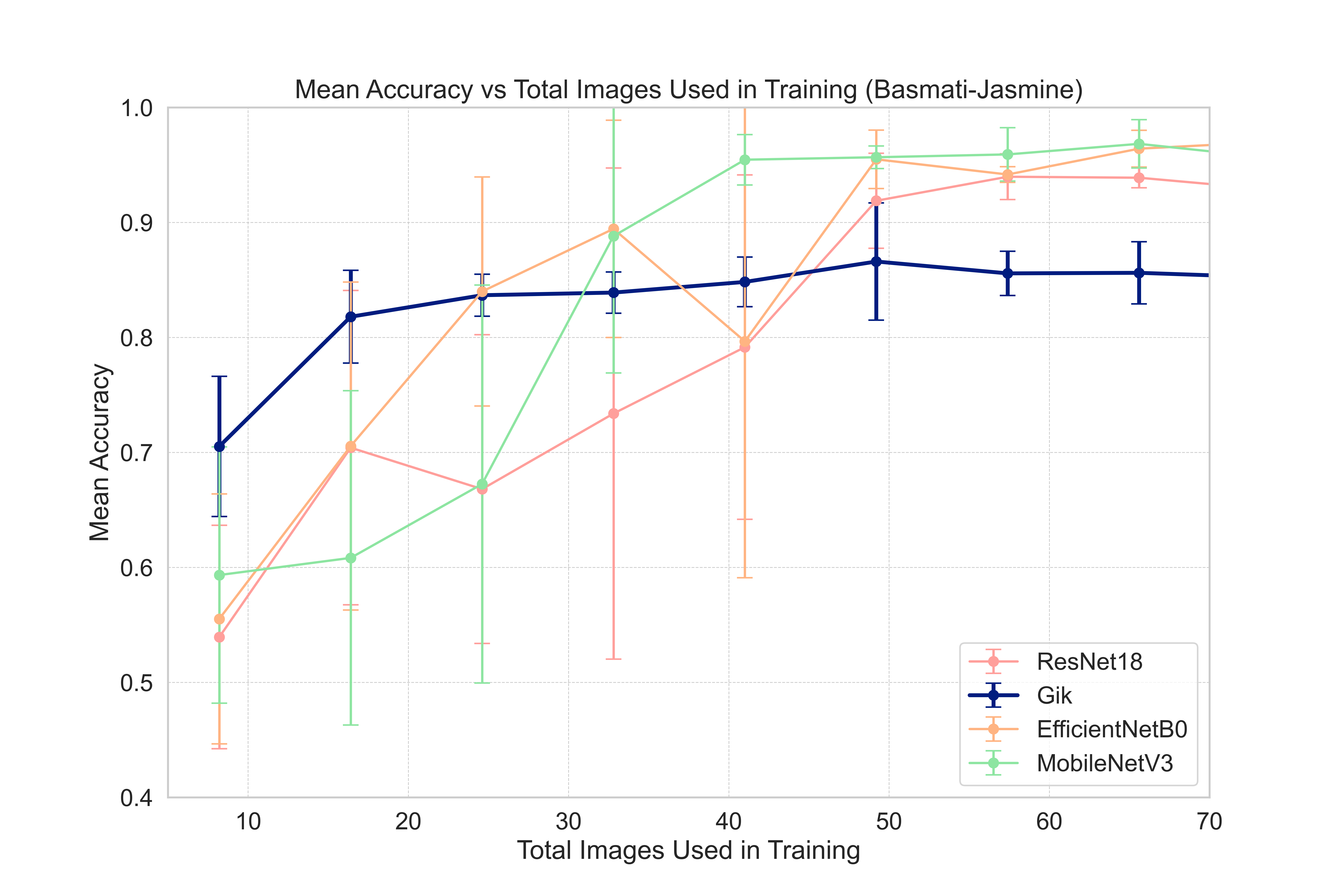}
    \caption{Mean accuracy in binary classification across the classes \textit{Karacadag} and \textit{Arborio}}
    \label{fig:KarcacadagAroborioGraph}
\end{figure*}

As illustrated in \textit{Figure \ref{fig:KarcacadagAroborioGraph}} (representing the Arborio-Karacadag dataset) and \textit{Figure \ref{fig:JasmineBasmatiGraph}} (representing the Basmati-Jasmine dataset), we contrast the performance of deep learning models against parameter-less models. The plotted accuracies represent the mean values derived from five training iterations, with the error bars indicating the corresponding standard deviations.

We verified with these results that in settings with a limited number of training examples parameter-less models, that exhibit some form of strong inductive bias, consistently outperform their deep learning counterparts. This observation aligns with the understanding that, due to this stronger inductive bias, parameter-less models are better equipped to distinguish between classes when given a reduced training set.  In this case, the Kolmogorov Complexity obtained by the Gik architecture aids in discriminating between images much faster than heavily trained models.

However, as the volume of training data expands, deep learning models start to surpass the parameter-less models. This can arguably be attributed to their ability to learn more comprehensive representations of the training distribution compared to what Gik can capture.

In \textit{Table \ref{tab:combined_performance_size_less_20}} performance and resource utilization is also expounded. This table compares the memory footprint, quantified in bytes, and mean accuracy for each model. It succinctly represents the effectiveness of Gik in achieving high accuracy with a very small memory footprint.

\begin{table*}
    \centering
    \begin{tabular}{l c c}
        \toprule
        Model & Accuracy (\%) & Model Size \\
        \midrule
        Gik  & \textbf{77.91} &\textbf{ 1.62 kB} \\
        EfficientNetB0 & 61.77 & 21.62 MB \\
        ResNet18 & 58.57 & 46.91 MB \\
        MobileNetV3 & 55.70 & 2.40 MB \\
        \bottomrule
    \end{tabular}
    \caption{Comparison of Model Performance and Size for Less than 20 Images Used}
    \label{tab:combined_performance_size_less_20}
\end{table*}

\section{Conclusion and Future Work}
\label{conclusion}

A central theme of this paper is the idea that how alternative methods which incorporate strong inductive bias can outperform with a reduced number of training samples other methods that try to learn from the data the insights already encoded in the inductive bias. 

Future work should be encouraged in terms of developing ways to formalize and measure the inductive bias presented in any machine learning models that derives inductive inference from data. We believe that such a formalization can have real-world applications in the selection process of machine learning models. Depending on the entropy of the problem distribution one can make use of a model with a stronger inductive bias, if the entropy is lower, or a weaker inductive bias, if the distribution's entropy is larger. The latter demanding more complex models, extensive datasets and compute than the former models.

In conclusion, we show that our model offers a computationally efficient alternative to mainstream deep learning models in a few-shot setting provided a high-quality dataset. Moreover, we have highlighted the potential behind architectures with pronounced inductive bias in terms of  accuracy and computational efficiency. Further research can be done on doing an ensemble of such models to achieve better results. We believe that architectures as the one presented can be especially fruitful for datasets where few images are available as well as memory constrained environments, like mobile or Cyber-Physical Systems.

\bibliographystyle{unsrtnat}
\bibliography{gik}  %%% Uncomment this line and comment out the ``thebibliography'' section below to use the external .bib file (using bibtex) .





% \end{thebibliography}

\end{document}